\documentclass[conference]{IEEEtran}
\IEEEoverridecommandlockouts
\usepackage{cite}
\usepackage{amsmath,amssymb,amsfonts}
\usepackage{algorithmic}
\usepackage{booktabs}
\usepackage{graphicx}
\usepackage{soul}
\usepackage{textcomp}
\usepackage{xcolor}
\def\BibTeX{{\rm B\kern-.05em{\sc i\kern-.025em b}\kern-.08em
    T\kern-.1667em\lower.7ex\hbox{E}\kern-.125emX}}
\begin{document}

\title{LRMR: LLM-Driven Relational Multi-node Ranking for Lymph Node Metastasis Assessment in Rectal Cancer\\
}

\author{
    \IEEEauthorblockN{
        Yaoxian Dong\textsuperscript{1,3},
        Yifan Gao\textsuperscript{1,3},
        Haoyue Li\textsuperscript{1,3},
        Yanfen Cui\textsuperscript{2}, and
        Xin Gao\textsuperscript{3*}
    }
    \IEEEauthorblockA{
        \textsuperscript{1}School of Biomedical Engineering (Suzhou), Division of Life Science and Medicine, \\
        University of Science and Technology of China, Hefei, China \\
    }
    \IEEEauthorblockA{
        \textsuperscript{2}Shanxi Province Cancer Hospital, Chinese Academy of Medical Sciences, Taiyuan, China \\
    }
    \IEEEauthorblockA{
        \textsuperscript{3}Suzhou Institute of Biomedical Engineering and Technology, Chinese Academy of Sciences, Suzhou, China
    }
    \IEEEauthorblockA{
        *Corresponding author 
    }
}

\maketitle

\begin{abstract}
Accurate preoperative assessment of lymph node (LN) metastasis in rectal cancer guides treatment decisions, yet conventional MRI evaluation based on morphological criteria shows limited diagnostic performance. While some artificial intelligence models have been developed, they often operate as black boxes, lacking the interpretability needed for clinical trust. Moreover, these models typically evaluate nodes in isolation, overlooking the patient-level context. To address these limitations, we introduce LRMR, an LLM-Driven Relational Multi-node Ranking framework. This approach reframes the diagnostic task from a direct classification problem into a structured reasoning and ranking process. The LRMR framework operates in two stages. First, a multimodal large language model (LLM) analyzes a composite montage image of all LNs from a patient, generating a structured report that details ten distinct radiological features. Second, a text-based LLM performs pairwise comparisons of these reports between different patients, establishing a relative risk ranking based on the severity and number of adverse features. We evaluated our method on a retrospective cohort of 117 rectal cancer patients. LRMR achieved an area under the curve (AUC) of 0.7917 and an F1-score of 0.7200, outperforming a range of deep learning baselines, including ResNet50 (AUC 0.7708). Ablation studies confirmed the value of our two main contributions: removing the relational ranking stage or the structured prompting stage led to a significant performance drop, with AUCs falling to 0.6875 and 0.6458, respectively. Our work demonstrates that decoupling visual perception from cognitive reasoning through a two-stage LLM framework offers a powerful, interpretable, and effective new paradigm for assessing lymph node metastasis in rectal cancer.
\end{abstract}


\section{Introduction}
The assessment of lymph node status is a cornerstone in the management of rectal cancer \cite{borgheresi2022lymph,ong2016assessment,fumero2011rim}. The presence or absence of nodal metastasis is a powerful prognostic indicator that heavily influences therapeutic strategies \cite{ji2023lymph}. An accurate preoperative N-staging distinguishes patients who may proceed directly to surgery from those who would benefit from neoadjuvant treatments \cite{delitto2018prognostic,maksim2024imaging}. The clinical impact is substantial, as the involvement of even a single lymph node can considerably alter a patient's expected survival rate. Therefore, precise staging before any intervention is of high importance for tailoring optimal, individualized treatment plans and avoiding both under-treatment and over-treatment \cite{tian2023deep}. Magnetic Resonance Imaging (MRI) has been established as the primary imaging modality for the local staging of rectal cancer \cite{ebaid2025comparing, horvat2019mri, beets2011local}.

Despite its central role, the current standard of MRI-based lymph node evaluation faces considerable challenges \cite{zhuang2021magnetic}. The diagnostic process has long depended on morphological features, with a pronounced reliance on size criteria. However, clinical evidence has repeatedly demonstrated the limitations of this approach. Malignant involvement can occur in nodes that are not enlarged, leading to false-negative results and potential disease recurrence \cite{yamamoto2022micrometastasis,zwart2022multimodal}. Conversely, nodal enlargement can be a result of benign reactive hyperplasia from inflammation or infection, producing false-positive findings that may lead to unnecessary invasive procedures. Other features, such as shape and border characteristics, also exhibit an overlap between benign and malignant conditions, which complicates diagnostic accuracy \cite{zhuang2021magnetic, rooney2022role, ozaki2025diagnostic}.

\begin{figure*}[h]
\centering
\includegraphics[width=\textwidth]{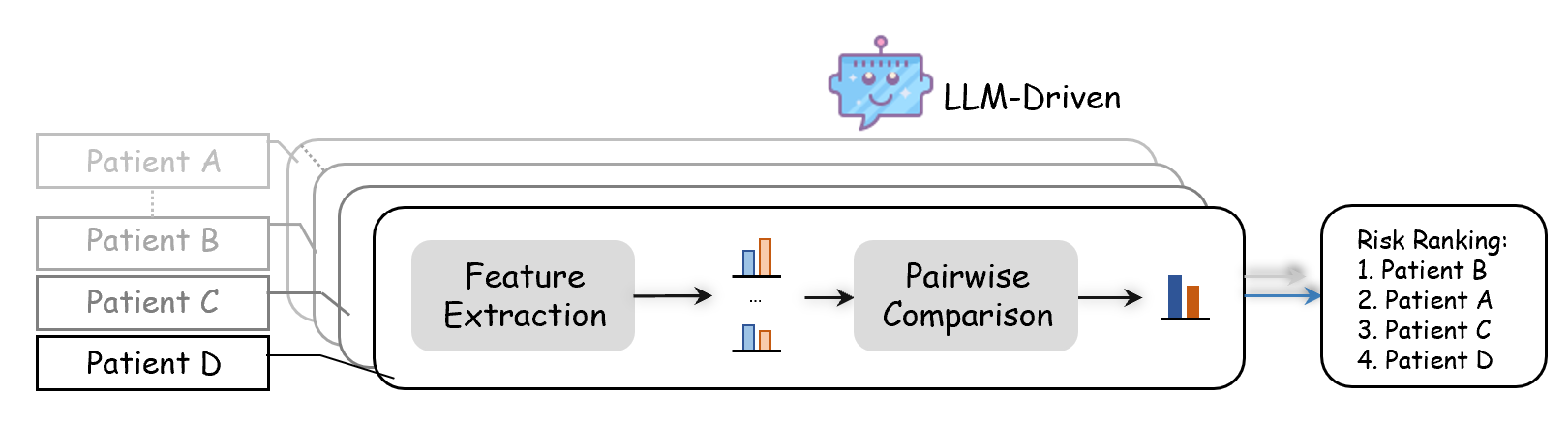}
\caption{The overall pipeline of the proposed framework. For each patient, all T2-weighted lymph node images are first compiled into a single montage image. In Stage 1, a multimodal LLM analyzes this montage to generate a structured JSON report detailing key radiological features. In Stage 2, a text-based LLM performs pairwise comparisons of these reports between different patients. The results are aggregated to produce a final risk score and ranking for each patient.}
\label{fig:pipeline}
\end{figure*}

To overcome the limitations of subjective human interpretation, artificial intelligence (AI), particularly deep learning, has emerged as a promising direction \cite{gao2023anatomy,wang2025prior,gao2024mba, dai2021transmed, gao2025safeclick}. These models have demonstrated strong performance in various medical image analysis tasks \cite{wu2025unsupervised,gao2024desam,xu2025tooth,chen2024emcnet}. However, many contemporary AI models for this purpose operate as black boxes, where the reasoning behind a prediction is not transparent. This lack of interpretability poses a significant barrier to their acceptance and trust in clinical practice \cite{liang2025multimodal}. Furthermore, most existing models evaluate each lymph node as an isolated entity \cite{xia2024multicenter,gao2025wega}. This approach neglects the holistic, patient-level perspective that clinicians often use, where the overall pattern and constellation of all visible nodes contribute to the final assessment.

In this paper, we propose the LLM-Driven Relational Multi-node Ranking (LRMR) framework, a new paradigm for assessing lymph node metastasis in rectal cancer. We reframe the problem from a direct classification task to a two-stage process of structured analysis and relational reasoning. In the first stage, a multimodal Large Language Model (LLM) examines a composite image of all lymph nodes from a single patient and generates a detailed, structured report on ten key radiological features. In the second stage, a text-based LLM compares these structured reports between different patients in a pairwise manner to establish a relative ranking of metastasis risk. This design allows for a comprehensive, patient-level assessment that is both effective and highly interpretable. Our main contributions include the novel two-stage framework, a montage-based approach for holistic multi-node analysis, and a relational ranking method based on LLM-driven comparison of structured data.

The remainder of this paper is organized as follows. Section 2 details our proposed methodology. Section 3 describes the experimental setup and presents the results, including comparisons with baseline models and ablation studies. Section 4 discusses the findings and concludes the paper.

\section{Methods}

\subsection{Overview}
Our proposed LRMR framework decomposes the complex task of lymph node metastasis assessment into two distinct stages: Structured Feature Extraction and Relational Risk Ranking. The overall pipeline of our approach is illustrated in \ref{fig:pipeline}. In the first stage, we address the challenge of holistic, patient-level evaluation. All lymph node images from a single patient are programmatically compiled into a single montage image, which is then analyzed by a multimodal LLM. This model functions as an automated reporting radiologist, generating a structured JSON report that details the status of ten prespecified radiological features across all visible nodes. The second stage leverages these structured reports for high-level cognitive reasoning. It employs a text-based LLM to perform pairwise comparisons between the reports of different patients. By evaluating the textual descriptions of adverse features, this model determines the relative risk of metastasis between two patients. The outcomes of these comparisons are then aggregated to compute a final risk score for each patient, establishing a global ranking. The specifics of each stage are detailed in the following subsections.

\begin{figure*}[t]
\centering
\includegraphics[width=\textwidth]{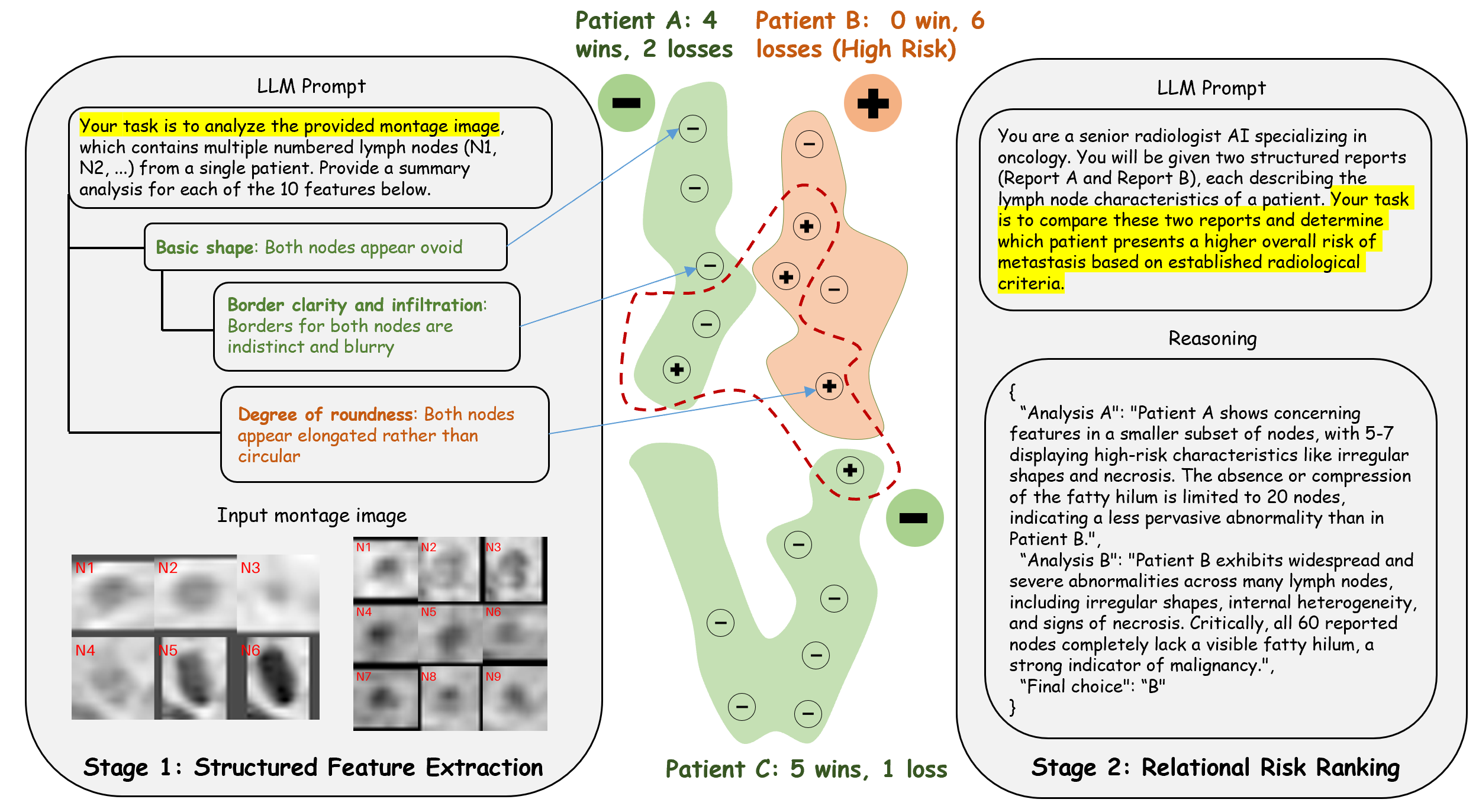}
\caption{Detailed illustration of the two-stage LRMR framework. (a) Stage 1: Structured Feature Extraction. All lymph node patches from a patient are compiled into a labeled montage image. This image, along with a structured textual prompt, is provided to a multimodal LLM, which outputs a detailed JSON report. (b) Stage 2: Relational Risk Ranking. The JSON reports from two different patients are inserted into a new prompt for a text-based LLM, which acts as an expert reasoner to compare the reports and determine the patient with the higher relative risk.}
\label{fig:method_detail}
\end{figure*}

\subsection{Structured Feature Extraction}
The initial stage of our LRMR framework is designed to overcome the limitations of isolated node analysis by translating raw visual data into a structured, patient-level report. The objective is to produce a comprehensive and machine-readable summary of all nodal findings for a single patient. The process begins with the programmatic compilation of all available T2-weighted lymph node image patches for a given patient. To enable a holistic assessment in one analytical step, these individual patches are algorithmically arranged into a single composite 'montage' image, as illustrated in \ref{fig:method_detail}a. During this compilation, each patch is resized to a uniform dimension of 128x128 pixels to standardize the input, and a unique alphanumeric label (e.g., N1, N2) is overlaid on the corner of each sub-image. This labeling provides a clear reference for the model to cite specific nodes in its analysis.

With the montage image prepared, we employ a multimodal LLM to perform a detailed visual examination. The model is guided by a carefully designed prompt that structures its analytical process. This prompt contains ten specific questions targeting clinically relevant radiological features, including lymph node shape, border contour and clarity, internal signal texture, and the presence or absence of key structures like the fatty hilum or areas of necrosis. A defining instruction within the prompt compels the model to not only summarize its findings for each feature but also to explicitly list the unique identifiers of any nodes that exhibit abnormal characteristics. The model is directed to output its entire analysis as a single, structured JSON object. The resulting report thereby serves as a high-level feature abstraction, converting complex pixel data from multiple nodes into a compact and rich textual representation of the patient's overall nodal status. This structured report forms the exclusive input for the subsequent reasoning stage, effectively decoupling the task of visual perception from cognitive analysis.

\subsection{Relational Risk Ranking}
The second stage of the LRMR framework uses the structured reports generated in the first stage to perform a high-level reasoning task, aiming to establish a robust risk ranking across the entire patient cohort. This stage operates exclusively in the text domain. We chose a relational ranking approach because determining a patient's risk profile by direct comparison against peers can be more robust than assigning an absolute, independent score. This method focuses on discerning relative differences in disease presentation, which is well-suited for handling the wide spectrum of findings in clinical practice. The process is built upon a pairwise comparison framework, as shown in \ref{fig:method_detail}b. For the entire patient cohort, we systematically generate comparison pairs. For each patient, we randomly sample a set of N opponents (where N=6 in our experiments) from the cohort to create a diverse set of comparison tasks.

For each generated pair (Patient A, Patient B), a new text-only prompt is dynamically constructed. This prompt frames the task for a text-based LLM, instructing it to act as a senior expert reviewing two radiological reports. The full JSON reports for both Patient A and Patient B, generated in Stage 1, are embedded directly into this prompt. The model's objective is to perform a cognitive comparison of the textual evidence. It must synthesize the information across all ten features for both patients, weighing factors such as the type, severity, and number of cited abnormalities. For instance, the model must reason whether a single node with extensive necrosis in Patient A constitutes a higher risk than three nodes with irregular borders in Patient B. The model's output is a structured JSON object containing its comparative analysis and a final choice ('A', 'B', or 'Comparable') indicating which patient presents a higher risk profile. After all comparisons are executed, the outcomes are aggregated. A final LRMR risk score for each patient is calculated by subtracting their total number of losses from their total number of wins ("Net Wins"). This single, continuous score reflects the patient's relative risk of metastasis within the cohort and is used for the final ranking and classification.

\begin{table*}[h]
\centering
\caption{Performance Comparison with Baseline Models on the Test Set. Best results are in \textbf{bold}, second-best are \underline{underlined}.}
\label{tab:main_results}
\begin{tabular}{llccccc}
\toprule
\textbf{Model} & \textbf{Type} & \textbf{Accuracy} & \textbf{Precision} & \textbf{Recall} & \textbf{F1-Score} & \textbf{AUC} \\
\midrule
DenseNet121 \cite{huang2017densely}     & CNN         & 0.6667          & 0.5000          & 0.7500          & 0.6000          & 0.6875 \\
EfficientNet-B0 \cite{tan2019efficientnet} & CNN         & 0.6944          & 0.5263          & \ul{0.8333}     & 0.6452          & 0.7292 \\
ResNet50 \cite{he2016deep}       & CNN         & \textbf{0.8056} & \textbf{0.7273} & 0.6667          & \ul{0.6957}     & \ul{0.7708} \\
ConvNeXt \cite{liu2022convnet}        & CNN         & 0.5278          & 0.4000          & \ul{0.8333}     & 0.5405          & 0.6042 \\
\midrule
ViT-Base \cite{dosovitskiy2020image}        & Transformer & 0.3889          & 0.3529          & \textbf{1.0000} & 0.5217          & 0.5417 \\
BEiT-Large \cite{bao2021beit}      & Transformer & 0.6389          & 0.4545          & 0.4167          & 0.4348          & 0.5833 \\
\midrule
MLP-Mixer \cite{tolstikhin2021mlp}       & MLP         & 0.6944          & 0.5455          & 0.5000          & 0.5217          & 0.6458 \\
ResMLP \cite{touvron2022resmlp}         & MLP         & 0.5278          & 0.3913          & 0.7500          & 0.5143          & 0.5833 \\
\midrule
MambaOut \cite{yu2025mambaout}           & Mamba       & \ul{0.7222}     & 0.5714          & 0.6667          & 0.6154          & 0.7083 \\
\midrule
\textbf{LRMR (Ours)} & Ours & \textbf{0.8056} & \ul{0.6923}     & 0.7500          & \textbf{0.7200} & \textbf{0.7917} \\
\bottomrule
\end{tabular}
\end{table*}

\section{Experiments and Results}

\subsection{Dataset}
This study was based on a retrospective cohort of 117 patients with pathologically confirmed rectal cancer from Fudan University Shanghai Cancer Center. All patients included in the study underwent surgical resection, and their definitive metastasis status was confirmed by postoperative histopathology, which provided a patient-level ground truth label (metastatic or non-metastatic). For each patient, preoperative pelvic MRI scans were acquired, from which T2-weighted imaging sequences were used for our analysis. The full patient cohort was divided into a training set and a test set using a 70/30 split. To ensure a comparable distribution of positive and negative cases in both sets, we employed stratified sampling based on the histopathological labels. This procedure resulted in a training set of 81 patients and a held-out test set of 36 patients. The training set was used exclusively to determine the optimal classification threshold from the LRMR risk scores, while the test set was used for the final, unbiased evaluation of model performance.

\begin{table*}[h]
\centering
\caption{Results of Ablation Studies on the Test Set.}
\label{tab:ablation_results}
\begin{tabular}{lccccc}
\toprule
\textbf{Model} & \textbf{Accuracy} & \textbf{Precision} & \textbf{Recall} & \textbf{F1-Score} & \textbf{AUC} \\
\midrule
LRMR (Full Model) & 0.8056 & 0.6923 & 0.7500 & 0.7200 & 0.7917 \\
LRMR w/o Relational Ranking & 0.6667 & 0.5000 & 0.7500 & 0.6000 & 0.6875 \\
LRMR w/o Structured Prompt & 0.6111 & 0.4500 & 0.7500 & 0.5625 & 0.6458 \\
\bottomrule
\end{tabular}
\end{table*}

\subsection{Experimental Setup}

The LRMR framework and all baseline models were implemented using Python 3.11. The deep learning baseline models were built and trained using the PyTorch library. The LLM used for both Stage 1 and Stage 2 was Google's gemini-2.5-flash-preview-05-20. For the multimodal LLM, the model's temperature was set to 0.1 to encourage more deterministic and factual visual descriptions.

In the Relational Risk Ranking stage, we generated pairwise comparisons across the entire patient cohort, with each patient being compared against 6 randomly selected opponents. The text-only prompt containing the two reports was then processed by the LLM to obtain a comparative risk judgment. For training the deep learning baselines, a patient-level feature representation was created by applying an aggregation function over the features extracted from all individual lymph node images of that patient. These models were trained using the AdamW optimizer with a cosine annealing learning rate scheduler for up to 50 epochs. All experiments were conducted on a NVIDIA RTX 4070Ti GPU.

To provide a comprehensive assessment of our model and the compared methods, we employed a set of standard evaluation metrics for binary classification tasks. The primary metric for evaluating the overall discriminative ability of the models across all possible thresholds was the Area Under the Receiver Operating Characteristic Curve (AUC). In addition, we calculated Accuracy to measure the proportion of correctly classified patients. To account for potential class imbalance and provide a more complete picture of the classification performance, we also report Precision, Recall (also known as Sensitivity), and the F1-score, which is the harmonic mean of Precision and Recall.

\subsection{Compared Methods}
We evaluated the performance of our proposed LRMR framework against a diverse array of established deep learning models to provide a thorough comparison. These baseline models were trained to predict the patient-level metastasis status directly from the aggregated lymph node image features. The baselines included widely-used Convolutional Neural Network (CNN) architectures such as ResNet50 \cite{he2016deep}, DenseNet121 \cite{huang2017densely}, and EfficientNet-B0 \cite{tan2019efficientnet}. We also included several more recent architectures to ensure a robust comparison, including the Vision Transformer (ViT-Base) \cite{dosovitskiy2020image}, ConvNeXt \cite{liu2022convnet}, BEiT-Large \cite{bao2021beit}, and models based on Multi-Layer Perceptrons like MLP-Mixer \cite{tolstikhin2021mlp} and ResMLP \cite{touvron2022resmlp}. Furthermore, we compared our method against MambaOut \cite{yu2025mambaout}, a recent model based on state-space architectures, to cover different families of deep learning models.

\subsection{Performance Comparison}
The main results comparing the LRMR framework with all baseline models on the held-out test set are presented in \ref{tab:main_results}. Our proposed LRMR model demonstrated superior performance across the majority of key evaluation metrics. It achieved an AUC of 0.7917, an F1-score of 0.7200, and an accuracy of 0.8056. This performance surpassed that of the strongest baseline model, ResNet50, which recorded an AUC of 0.7708 and an F1-score of 0.6957. Other competitive models like EfficientNet-B0 and MambaOut achieved AUCs of 0.7292 and 0.7083, respectively. In contrast, several other architectures showed lower performance. For example, ViT-Base obtained a recall of 1.0000 but a very low precision of 0.3529, indicating a model bias towards positive predictions that lacked effective discriminative capability, as reflected by its low AUC of 0.5417. These results collectively show that the two-stage reasoning process of LRMR is more effective for this assessment task than direct end-to-end classification with various standard deep learning models.

\subsection{Ablation Study}
To validate the contributions of the primary components of our framework, we conducted two ablation studies. The results are summarized in \ref{tab:ablation_results}. In the first experiment, titled 'LRMR w/o Relational Ranking', we removed the second-stage relational comparison LLM. Instead, a simple rule-based scoring method was applied directly to the JSON reports from Stage 1. This change resulted in a sharp drop in performance, with the AUC decreasing from 0.7917 to 0.6875 and the F1-score falling from 0.7200 to 0.6000. This finding confirms the importance of the relational reasoning stage, suggesting that an LLM-driven comparison of structured reports provides a more nuanced and accurate risk assessment than a direct aggregation of feature scores.

In the second experiment, 'LRMR w/o Structured Prompt', we replaced the detailed 10-question prompt in Stage 1 with a simple, open-ended request for a free-form risk description. This modification led to an even more pronounced degradation in performance, with the AUC decreasing to 0.6458 and the F1-score to 0.5625. The performance of this ablated model was lower than most of the deep learning baselines. This result underscores the value of our structured prompting mechanism. Guiding the LLM's analysis with a checklist of specific, clinically relevant features is necessary for extracting reliable and consistent information. Without this structured guidance, the feature extraction process becomes unstable and less effective, which consequently undermines the performance of the entire framework.

\section{Discussion and Conclusion}
In this study, we introduced and evaluated LRMR, a two-stage framework that leverages Large Language Models for the assessment of lymph node metastasis in rectal cancer. Our experimental results show that the LRMR approach achieved a high level of performance, obtaining an AUC of 0.7917 and an F1-score of 0.7200 on the test set. This performance surpassed that of a wide range of standard deep learning models, including strong baselines like ResNet50. The success of our method is not merely a numerical improvement but a validation of our core hypothesis: decoupling the complex diagnostic task into a visual perception stage and a subsequent cognitive reasoning stage is a more effective strategy than direct end-to-end classification. The ablation studies provide direct support for this conclusion. The removal of either the structured prompting mechanism in Stage 1 or the relational ranking process in Stage 2 resulted in a considerable drop in performance, as detailed in \ref{tab:ablation_results}. This confirms that both the quality of the initial feature abstraction and the sophistication of the comparative reasoning are integral to the framework's success.

A primary motivation for this work was to address the black box nature of many contemporary AI models. The LRMR framework offers a transparent and interpretable alternative. The entire decision-making process is auditable. The structured JSON report from Stage 1 provides a clear account of the specific radiological features observed in each node, while the output from Stage 2 offers an explicit textual justification for why one patient was deemed to have a higher risk than another. This moves beyond providing a simple probability score and instead delivers a reasoned assessment that is more aligned with the diagnostic workflows of clinicians. By structuring the task in this way, our framework represents a step towards developing AI systems that do not just recognize patterns but also perform a form of automated clinical reasoning.

Furthermore, our approach addresses the limitation of evaluating lymph nodes in isolation. By compiling all of a patient's nodes into a single montage for analysis, the Stage 1 LLM is able to form a holistic, patient-level impression. The relational ranking performed in Stage 2 provides a more nuanced output than a binary classification. A patient's final risk score is not an absolute value but is defined relative to their peers in the cohort. This context-aware ranking can be particularly valuable in clinical practice for tasks such as patient triage or for identifying borderline cases that may require closer follow-up or further diagnostic workup.

In conclusion, we presented LRMR, a novel two-stage, LLM-driven framework that reframes lymph node assessment as a process of structured reporting and relational reasoning. Our results show that this approach is not only effective, outperforming various deep learning baselines, but is also highly interpretable. By separating visual analysis from cognitive comparison and by performing a holistic, multi-node assessment, our work offers a promising new direction for the application of artificial intelligence in radiological diagnostics, moving beyond simple classification towards more human-aligned reasoning tasks.

\bibliographystyle{IEEEtran}
\bibliography{conf.bib}

\begin{thebibliography}{10}
\providecommand{\url}[1]{#1}
\csname url@samestyle\endcsname
\providecommand{\newblock}{\relax}
\providecommand{\bibinfo}[2]{#2}
\providecommand{\BIBentrySTDinterwordspacing}{\spaceskip=0pt\relax}
\providecommand{\BIBentryALTinterwordstretchfactor}{4}
\providecommand{\BIBentryALTinterwordspacing}{\spaceskip=\fontdimen2\font plus
\BIBentryALTinterwordstretchfactor\fontdimen3\font minus \fontdimen4\font\relax}
\providecommand{\BIBforeignlanguage}[2]{{%
\expandafter\ifx\csname l@#1\endcsname\relax
\typeout{** WARNING: IEEEtran.bst: No hyphenation pattern has been}%
\typeout{** loaded for the language `#1'. Using the pattern for}%
\typeout{** the default language instead.}%
\else
\language=\csname l@#1\endcsname
\fi
#2}}
\providecommand{\BIBdecl}{\relax}
\BIBdecl

\bibitem{borgheresi2022lymph}
A.~Borgheresi, F.~De~Muzio, A.~Agostini, L.~Ottaviani, A.~Bruno, V.~Granata, R.~Fusco, G.~Danti, F.~Flammia, R.~Grassi \emph{et~al.}, ``Lymph nodes evaluation in rectal cancer: where do we stand and future perspective,'' \emph{Journal of Clinical Medicine}, vol.~11, no.~9, p. 2599, 2022.

\bibitem{ong2016assessment}
M.~L. Ong and J.~B. Schofield, ``Assessment of lymph node involvement in colorectal cancer,'' \emph{World journal of gastrointestinal surgery}, vol.~8, no.~3, p. 179, 2016.

\bibitem{fumero2011rim}
F.~Fumero, S.~Alay{\'o}n, J.~L. Sanchez, J.~Sigut, and M.~Gonzalez-Hernandez, ``Rim-one: An open retinal image database for optic nerve evaluation,'' in \emph{2011 24th international symposium on computer-based medical systems (CBMS)}.\hskip 1em plus 0.5em minus 0.4em\relax IEEE, 2011, pp. 1--6.

\bibitem{ji2023lymph}
H.~Ji, C.~Hu, X.~Yang, Y.~Liu, G.~Ji, S.~Ge, X.~Wang, and M.~Wang, ``Lymph node metastasis in cancer progression: molecular mechanisms, clinical significance and therapeutic interventions,'' \emph{Signal Transduction and Targeted Therapy}, vol.~8, no.~1, p. 367, 2023.

\bibitem{delitto2018prognostic}
D.~Delitto, T.~J. George~Jr, T.~J. Loftus, P.~Qiu, G.~J. Chang, C.~J. Allegra, W.~A. Hall, S.~J. Hughes, S.~A. Tan, C.~M. Shaw \emph{et~al.}, ``Prognostic value of clinical vs pathologic stage in rectal cancer patients receiving neoadjuvant therapy,'' \emph{JNCI: Journal of the National Cancer Institute}, vol. 110, no.~5, pp. 460--466, 2018.

\bibitem{maksim2024imaging}
R.~Maksim, A.~Buczy{\'n}ska, I.~Sidorkiewicz, A.~J. Kretowski, and E.~Sierko, ``Imaging and metabolic diagnostic methods in the stage assessment of rectal cancer,'' \emph{Cancers}, vol.~16, no.~14, p. 2553, 2024.

\bibitem{tian2023deep}
C.~Tian, X.~Ma, H.~Lu, Q.~Wang, C.~Shao, Y.~Yuan, and F.~Shen, ``Deep learning models for preoperative t-stage assessment in rectal cancer using mri: exploring the impact of rectal filling,'' \emph{Frontiers in Medicine}, vol.~10, p. 1326324, 2023.

\bibitem{ebaid2025comparing}
N.~Y. Ebaid, S.~E. Badr, R.~F. Mansour, H.~A. Abo-Alella, M.~M. Assy, S.~K.~S. Eldemerdash, M.~A. S.~A. Haasan, and H.~A.~E. Mohamed, ``Comparing abbreviated and full mri protocols for preoperative local staging of locally advanced rectal cancer,'' \emph{Academic Radiology}, 2025.

\bibitem{horvat2019mri}
N.~Horvat, C.~Carlos Tavares~Rocha, B.~Clemente~Oliveira, I.~Petkovska, and M.~J. Gollub, ``Mri of rectal cancer: tumor staging, imaging techniques, and management,'' \emph{Radiographics}, vol.~39, no.~2, pp. 367--387, 2019.

\bibitem{beets2011local}
R.~G. Beets-Tan and G.~L. Beets, ``Local staging of rectal cancer: a review of imaging,'' \emph{Journal of Magnetic Resonance Imaging}, vol.~33, no.~5, pp. 1012--1019, 2011.

\bibitem{zhuang2021magnetic}
Z.~Zhuang, Y.~Zhang, M.~Wei, X.~Yang, and Z.~Wang, ``Magnetic resonance imaging evaluation of the accuracy of various lymph node staging criteria in rectal cancer: a systematic review and meta-analysis,'' \emph{Frontiers in Oncology}, vol.~11, p. 709070, 2021.

\bibitem{yamamoto2022micrometastasis}
H.~Yamamoto, ``Micrometastasis in lymph nodes of colorectal cancer,'' \emph{Annals of Gastroenterological Surgery}, vol.~6, no.~4, pp. 466--473, 2022.

\bibitem{zwart2022multimodal}
W.~H. Zwart, A.~Hotca, G.~A. Hospers, K.~A. Goodman, and J.~Garcia-Aguilar, ``The multimodal management of locally advanced rectal cancer: making sense of the new data,'' \emph{American Society of Clinical Oncology Educational Book}, vol.~42, pp. 1--14, 2022.

\bibitem{rooney2022role}
S.~Rooney, J.~Meyer, Z.~Afzal, J.~Ashcroft, H.~Cheow, K.~DePaepe, M.~Powar, C.~Simillis, J.~Wheeler, J.~Davies \emph{et~al.}, ``The role of preoperative imaging in the detection of lateral lymph node metastases in rectal cancer: a systematic review and diagnostic test meta-analysis,'' \emph{Diseases of the Colon \& Rectum}, vol.~65, no.~12, pp. 1436--1446, 2022.

\bibitem{ozaki2025diagnostic}
K.~Ozaki, K.~Kawai, S.~Ogawa, Y.~Kanemitsu, and Y.~Ajioka, ``Diagnostic accuracy of lateral lymph node metastasis for locally advanced rectal cancer after neoadjuvant therapy: a systematic review and meta-analysis,'' \emph{Expert Review of Anticancer Therapy}, pp. 1--7, 2025.

\bibitem{gao2023anatomy}
Y.~Gao, Y.~Dai, F.~Liu, W.~Chen, and L.~Shi, ``An anatomy-aware framework for automatic segmentation of parotid tumor from multimodal mri,'' \emph{Computers in Biology and Medicine}, vol. 161, p. 107000, 2023.

\bibitem{wang2025prior}
T.~Wang, Y.~Gao, B.~Liang, and Q.~Wang, ``Prior-driven refinement network for small organ-at-risk segmentation in head and neck cancer,'' \emph{Engineering Applications of Artificial Intelligence}, vol. 159, p. 111605, 2025.

\bibitem{gao2024mba}
Y.~Gao, W.~Xia, W.~Wang, and X.~Gao, ``Mba-net: Sam-driven bidirectional aggregation network for ovarian tumor segmentation,'' in \emph{International Conference on Medical Image Computing and Computer-Assisted Intervention}.\hskip 1em plus 0.5em minus 0.4em\relax Springer, 2024, pp. 437--447.

\bibitem{dai2021transmed}
Y.~Dai, Y.~Gao, and F.~Liu, ``Transmed: Transformers advance multi-modal medical image classification,'' \emph{Diagnostics}, vol.~11, no.~8, p. 1384, 2021.

\bibitem{gao2025safeclick}
Y.~Gao, J.~Sheng, W.~Wu, H.~Li, Y.~Dong, C.~Ge, F.~Yuan, and X.~Gao, ``Safeclick: Error-tolerant interactive segmentation of any medical volumes via hierarchical expert consensus,'' \emph{arXiv preprint arXiv:2506.18404}, 2025.

\bibitem{wu2025unsupervised}
W.~Wu, Y.~Gao, X.~Jin, R.~Zhang, Y.~Pan, and X.~Gao, ``An unsupervised anatomy-aware dual-constraint cascade network for lung computed tomography deformable image registration,'' \emph{Engineering Applications of Artificial Intelligence}, vol. 158, p. 111548, 2025.

\bibitem{gao2024desam}
Y.~Gao, W.~Xia, D.~Hu, W.~Wang, and X.~Gao, ``Desam: Decoupled segment anything model for generalizable medical image segmentation,'' in \emph{International Conference on Medical Image Computing and Computer-Assisted Intervention}.\hskip 1em plus 0.5em minus 0.4em\relax Springer, 2024, pp. 509--519.

\bibitem{xu2025tooth}
X.~Xu, J.~Chen, and J.~Yin, ``Tooth instance segmentation and disease detection with uncertainty-aware contrastive learning and cross-scale attention,'' \emph{IEEE Journal of Biomedical and Health Informatics}, 2025.

\bibitem{chen2024emcnet}
J.~Chen, B.~Fang, H.~Li, L.-B. Zhang, Y.~Teng, and G.~Fortino, ``Emcnet: Ensemble multiscale convolutional neural network for single-lead ecg classification in wearable devices,'' \emph{IEEE Sensors Journal}, vol.~24, no.~6, pp. 8754--8762, 2024.

\bibitem{liang2025multimodal}
B.~Liang, Y.~Gao, T.~Wang, L.~Zhang, and Q.~Wang, ``Multimodal large language models address clinical queries in laryngeal cancer surgery: a comparative evaluation of image interpretation across different models,'' \emph{International Journal of Surgery}, vol. 111, no.~3, pp. 2727--2730, 2025.

\bibitem{xia2024multicenter}
W.~Xia, D.~Li, W.~He, P.~J. Pickhardt, J.~Jian, R.~Zhang, J.~Zhang, R.~Song, T.~Tong, X.~Yang \emph{et~al.}, ``Multicenter evaluation of a weakly supervised deep learning model for lymph node diagnosis in rectal cancer at mri,'' \emph{Radiology: Artificial Intelligence}, vol.~6, no.~2, p. e230152, 2024.

\bibitem{gao2025wega}
Y.~Gao, Y.~Dong, W.~Wu, C.~Ge, F.~Yuan, J.~Sheng, H.~Li, and X.~Gao, ``Wega: Weakly-supervised global-local affinity learning framework for lymph node metastasis prediction in rectal cancer,'' \emph{arXiv preprint arXiv:2505.10502}, 2025.

\bibitem{huang2017densely}
G.~Huang, Z.~Liu, L.~Van Der~Maaten, and K.~Q. Weinberger, ``Densely connected convolutional networks,'' in \emph{Proceedings of the IEEE conference on computer vision and pattern recognition}, 2017, pp. 4700--4708.

\bibitem{tan2019efficientnet}
M.~Tan and Q.~Le, ``Efficientnet: Rethinking model scaling for convolutional neural networks,'' in \emph{International conference on machine learning}.\hskip 1em plus 0.5em minus 0.4em\relax PMLR, 2019, pp. 6105--6114.

\bibitem{he2016deep}
K.~He, X.~Zhang, S.~Ren, and J.~Sun, ``Deep residual learning for image recognition,'' in \emph{Proceedings of the IEEE conference on computer vision and pattern recognition}, 2016, pp. 770--778.

\bibitem{liu2022convnet}
Z.~Liu, H.~Mao, C.-Y. Wu, C.~Feichtenhofer, T.~Darrell, and S.~Xie, ``A convnet for the 2020s,'' in \emph{Proceedings of the IEEE/CVF conference on computer vision and pattern recognition}, 2022, pp. 11\,976--11\,986.

\bibitem{dosovitskiy2020image}
A.~Dosovitskiy, L.~Beyer, A.~Kolesnikov, D.~Weissenborn, X.~Zhai, T.~Unterthiner, M.~Dehghani, M.~Minderer, G.~Heigold, S.~Gelly \emph{et~al.}, ``An image is worth 16x16 words: Transformers for image recognition at scale,'' \emph{arXiv preprint arXiv:2010.11929}, 2020.

\bibitem{bao2021beit}
H.~Bao, L.~Dong, S.~Piao, and F.~Wei, ``Beit: Bert pre-training of image transformers,'' \emph{arXiv preprint arXiv:2106.08254}, 2021.

\bibitem{tolstikhin2021mlp}
I.~O. Tolstikhin, N.~Houlsby, A.~Kolesnikov, L.~Beyer, X.~Zhai, T.~Unterthiner, J.~Yung, A.~Steiner, D.~Keysers, J.~Uszkoreit \emph{et~al.}, ``Mlp-mixer: An all-mlp architecture for vision,'' \emph{Advances in neural information processing systems}, vol.~34, pp. 24\,261--24\,272, 2021.

\bibitem{touvron2022resmlp}
H.~Touvron, P.~Bojanowski, M.~Caron, M.~Cord, A.~El-Nouby, E.~Grave, G.~Izacard, A.~Joulin, G.~Synnaeve, J.~Verbeek \emph{et~al.}, ``Resmlp: Feedforward networks for image classification with data-efficient training,'' \emph{IEEE transactions on pattern analysis and machine intelligence}, vol.~45, no.~4, pp. 5314--5321, 2022.

\bibitem{yu2025mambaout}
W.~Yu and X.~Wang, ``Mambaout: Do we really need mamba for vision?'' in \emph{Proceedings of the Computer Vision and Pattern Recognition Conference}, 2025, pp. 4484--4496.

\end{thebibliography}

\end{document}